\documentclass[10pt, conference, compsocconf]{IEEEtran}
\ifCLASSINFOpdf
\else
\fi
\usepackage[utf8]{inputenc}
\usepackage{threeparttable}
\usepackage{amssymb,adjustbox}
\usepackage{url}
\usepackage{array}
\usepackage{graphicx}
\usepackage{epsfig}
\usepackage{multirow}
\usepackage{makecell}
\usepackage{figsize}
\usepackage{CJKutf8} 
\usepackage{ucs} 
\usepackage[encapsulated]{CJK}

\begin{document}
%
\title{ICDAR 2019 Competition on Large-scale Street View Text with Partial Labeling - RRC-LSVT}


\author{\IEEEauthorblockN{Authors Name/s per 1st Affiliation (Author)}
\IEEEauthorblockA{line 1 (of Affiliation): dept. name of organization\\
line 2: name of organization, acronyms acceptable\\
line 3: City, Country\\
line 4: Email: name@xyz.com}
\and
\IEEEauthorblockN{Authors Name/s per 2nd Affiliation (Author)}
\IEEEauthorblockA{line 1 (of Affiliation): dept. name of organization\\
line 2: name of organization, acronyms acceptable\\
line 3: City, Country\\
line 4: Email: name@xyz.com}
}

\author{\IEEEauthorblockN{Yipeng Sun\IEEEauthorrefmark{1},  Zihan Ni\IEEEauthorrefmark{1}, Chee-Kheng Chng\IEEEauthorrefmark{2}, Yuliang Liu\IEEEauthorrefmark{3}, Canjie Luo\IEEEauthorrefmark{3}, Chun Chet Ng\IEEEauthorrefmark{2}, \\
Junyu Han\IEEEauthorrefmark{1}, Errui Ding\IEEEauthorrefmark{1}, Jingtuo Liu\IEEEauthorrefmark{1}, Dimosthenis Karatzas\IEEEauthorrefmark{4}, Chee Seng Chan\IEEEauthorrefmark{2}, Lianwen Jin\IEEEauthorrefmark{3}\\
\IEEEauthorrefmark{1}Baidu Inc, Beijing, China\\
\IEEEauthorrefmark{2}University of Malaya, Malaysia\\
\IEEEauthorrefmark{3}South China university of Technology, Guangdong, China\\
\IEEEauthorrefmark{4}Computer Vision Center, Universitat Aut\`onoma de Barcelona, Spain}
}



\maketitle

\begin{abstract}
Robust text reading from street view images provides valuable information for various applications. Performance improvement of existing methods in such a challenging scenario heavily relies on the amount of fully annotated training data, which is costly and in-efficient to obtain. To scale up the amount of training data while keeping the labeling procedure cost-effective, this competition introduces a new challenge on \bf{L}arge-scale \bf{S}treet \bf{V}iew \bf{T}ext with \bf{P}artial \bf{L}abeling (LSVT), providing $50,000$ and $400,000$ images in full and weak annotations, respectively. This competition aims to explore the abilities of state-of-the-art methods to detect and recognize text instances from large-scale street view images, closing the gap between research benchmarks and real applications. During the competition period, a total of $41$ teams participated in the two proposed tasks with $132$ valid submissions, i.e., text detection and end-to-end text spotting. This paper includes dataset descriptions, task definitions, evaluation protocols and results summaries of the ICDAR 2019-LSVT challenge.
\end{abstract}


\IEEEpeerreviewmaketitle

\section{INTRODUCTION}
Recent powerful deep learning models contributed dramatically to the advances of robust text reading problems, including text detection, recognition and end-to-end text spotting. Benefiting from the pioneer work of the existing benchmarks~\cite{yao2012detecting,karatzas13icdar,karatzas15icdar,liu2011casia,icdar2017-mlt,chng17tt,yuliang2017detecting,shi2017icdar2017,yuan2018chinese}, remarkable success has been achieved in text detection and recognition in the wild. 
Since most of the scene text datasets provide fully annotated ground truth (i.e. all ground truth for the problem is given), due to the heavy cost of full labelling, the amount of training data is relatively limited to optimize deep learning algorithms. Most of recent scene text understanding models \cite{jaderberg2014deep, he2017deep, xiangcvpr2017, TextBoxes, DMPNet, Liao2018TextBoxesAS, yuliang2017detecting} depend on pre-training on the 800,000-image synthetic dataset~\cite{Gupta16} and fine-tuning on real training data, which has become a common way in the field. The fact is that large-scale images and full annotations are expensive and inefficient to obtain. Even though the number of synthetic data is huge, there is still noticeable differences between synthetic and real data samples. Due to these reasons, incessant efforts on models with reliance on more fully annotated data should not be the only direction to advance in, for the field of robust text reading.

In this competition, we collected a new large-scale scene text dataset, namely Large-scale Street View Text with Partial Labeling (\textit{LSVT}), with $30,000$ training images in full annotations and $400,000$ training samples in weak annotations, which are referred to as partial labels. We intend to challenge the community to look into novel solutions which can further boost the performance from partial labels. For most of the training data in weak labels, only one transcription per image is provided without location annotations, which is referred to as `text-of-interest'. It is labeled as one of the keywords with semantic information, e.g., the name of a store front. In total, there are $450,000$ images in ICDAR 2019-LSVT including $20,000$ testing images, making it the largest scene text dataset to-date. All the images were captured from streets, which consist of a large variety of complicated real world scenarios, making the challenge extremely high by narrowing gaps between research and real applications.

To the best of our knowledge, ICDAR 2019-LSVT is at least $14$ times as large as existing robust reading benchmarks, and is also the first ever scene text dataset labeled with partial annotations for the text detection and recognition challenges. The amount of fully annotated part of data is also greater than that of previous robust reading benchmarks. This competition is an extension to the previous RRC competitions, specifically the tasks of scene text detection and recognition in natural images. The key aspects of the proposed dataset are:  i) fully annotated data and large-scale weakly annotated data in the training set, ii) the wild texts captured in the streets, and iii) the largest testing benchmark. Submitted models are required to have high generalization and robustness in order to strive in this competition.

\section{COMPETITION ORGANIZATION}
ICDAR 2019-LSVT competition is organized by a joint team of Baidu Inc., South China University of Technology, University of Malaya and the Computer Vision Centre (Universitat Aut\`onoma de Barcelona).
The competition started and the training data was released on Mar. 1. The first and second parts of the test set were released on Apr. 10 and 20, respectively. The submission entries opened on Apr. 20 and closed on Apr. 30, 23:59 PDT~(Pacific Daylight Time). Overall, we received $132$ valid submissions from $41$ teams from both research communities and industries for the two tasks.
\begin{figure*}[!htb]
    \centering
    \begin{minipage}{0.327\linewidth}
	\begin{center}
		\subfigure[]{\includegraphics[width=\linewidth]{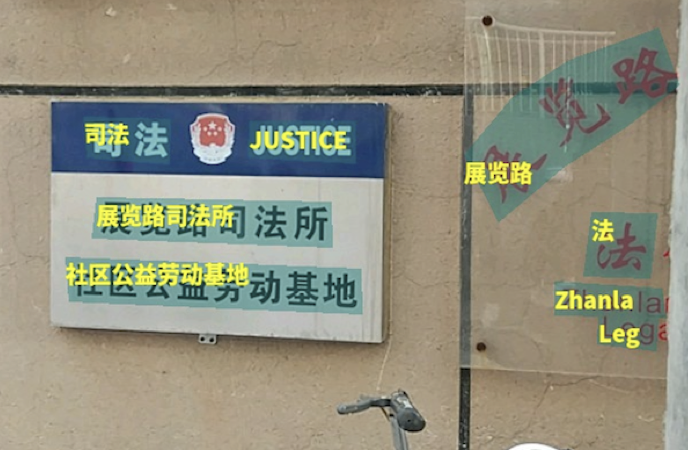}}
		\centering
	\end{center}
     \end{minipage}
     \begin{minipage}{0.327\linewidth}
	\begin{center}
		\subfigure[]{\includegraphics[width=\linewidth]{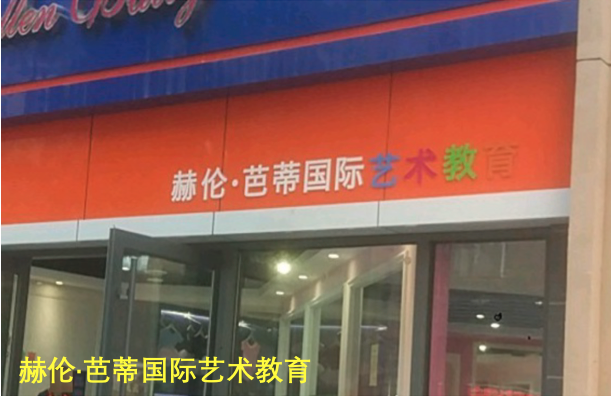}}
		\centering
	\end{center}
     \end{minipage}
     \begin{minipage}{0.327\linewidth}
	\begin{center}
		\subfigure[]{\includegraphics[width=\linewidth]{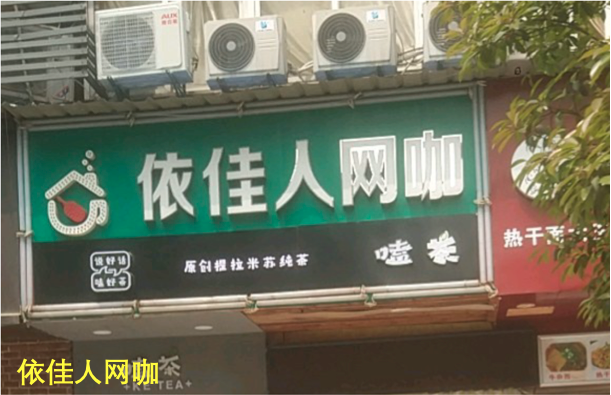}}
		\centering
	\end{center}
     \end{minipage}\vspace{-0.5em}
\caption{Examples of cropped from ICDAR 2019-LSVT dataset in full and weak annotations. For the fully annotated samples, the ground truth locations and corresponding text are labeled as shown in (a). For the weakly annotated samples, important keywords in these images, e.g., the name of a store font, are given as the ground truth without locations as shown in (b) and (c), which are much cheaper to collect and annotate. The ground truth of (b) and (c) are shown on the left bottom of figures, which indicate the keywords carrying meaningful information in these images.}
\label{fig:examples_full_weak_label}
\end{figure*}



\section{DATASETS}
\textbf{Type/source of images}
The dataset used for this competition comprises $450,000$ images with text that are freely captured in the streets of China, e.g., store fronts and landmarks. All images are captured by different users with different mobile phones. To address privacy issues, regions including faces and license plates are all detected by algorithms and blurred.

\textbf{Number of images}
There are $450,000$ images in the ICDAR 2019-LSVT dataset. $50,000$ of them are fully annotated as shown in Fig.~\ref{fig:examples_full_weak_label}(a), which are split into a training set of $30,000$ and a test set of $20,000$. The rest of the $400,000$ images are weakly annotated as shown in Fig.~\ref{fig:examples_full_weak_label} (b) and (c). During the testing stage of the competition, the test set is divided into two parts in half to release. To the best of our knowledge, the dataset is more than $14$ times as large as existing robust reading benchmarks, and it is the largest dataset on Chinese text to date. It is also the first scene text dataset labeled with partial annotations in ICDAR text detection and recognition challenges.

\textbf{Annotations}
The spatial coordinates and transcriptions of every text instance are annotated manually. For the fully annotated images in the dataset, horizontal, multi-oriented, and vertical text instances are labeled with quadrilateral bounding boxes, while curved text instances are annotated in polygons with $8$ or $12$ vertices. The points in the polygons are arranged in clockwise sequence, starting from reading direction. The transcriptions of every text instance are annotated and encoded in UTF-8. The illegibility regions in images are labeled as \emph{Do Not Care}, which are ignored to calculate final scores. For the weakly annotated images, only the transcriptions of the keywords are provided, which are referred to as weak annotations.

\section{TASKS}
To evaluate text reading performance, we introduce two common tasks on this large-scale street view text benchmarks, i.e., text detection and end-to-end text spotting.
\subsection{TASK 1 - TEXT DETECTION}
This task aims to tell the locations of text instances from images in bounding boxes or polygons.
Following the evaluation protocols of ICDAR 2015~\cite{karatzas15icdar}, the text detection task of ICDAR 2019-LSVT is evaluated in terms of Precision, Recall and H-mean with the IoU~(Intersection-over-Union) threshold of $0.5$ and $0.7$, while only H-mean (IoU higher than $0.5$) is regarded as the final metric for ranking. A detected text line is considered as true positive if the detected region has more than $0.5$ IoU with the ground truth box. Meanwhile, in the case of multiple matches, we only consider the detection region with the highest IoU, and the rest of the matches is counted as False Positive. Similar to COCO-text~\cite{veit2016coco} and ICDAR 2015~\cite{karatzas15icdar}, all detected or missed \emph{Do not care} ground truths do not contribute to the evaluation result. Precision, Recall, and H-mean are calculated as
\begin{equation}
Precision = \frac{TP}{TP + FP}, 
\end{equation}
\begin{equation}
Recall = \frac{TP}{TP + FN},
\end{equation}
and
\begin{equation}
Hmean = \frac{2*Precision*Recall}{Precision + Recall},
\end{equation}
where $TP$, $FP$ and $FN$ denote true positive, false positive and false negative, respectively. \\

\begin{table*}[t]
\caption{Text detection results and rankings after removing the duplicate submissions in Task 1. Note that * denotes missing descriptions in affiliations.} \vspace{-0.5em}
\label{task1_top_10}
\centering
\begin{tabular}
{p{3.3cm}<{\centering}|p{1.5cm}<{\centering}|p{1cm}<{\centering}|p{1cm}<{\centering}|p{0.8cm}<{\centering}|p{1cm}<{\centering}|p{6cm}<{\centering}}
\hline
\textbf{Team name} & \textbf{User ID} & \textbf{Rank} & \textbf{H-mean} & \textbf{Recall} & \textbf{Precision} & \textbf{Affiliations}  \\
\hline
\hline
Tencent-DPPR Team & user\_9062 & 1 & 86.42 & 83.31 & 89.77 & Tencent-DPPR team\\
\hline
NJU\_ImagineLab & user\_7597 & 2  & 85.61 & 82.09 & 89.45 & ImagineLab, National Key Lab for Novel Software Technology, Nanjing University\\
\hline
PMTD & user\_3056 & 3  & 84.59 & 81.92 & 87.44 & SenseTime\\
\hline
DMText\_lsvt & user\_3237 & 4  & 84.40 & 80.03 & 89.28 & Tencent\\
\hline
baseline\_polygon\_0.7 &  user\_9007 & 5  & 84.37 & 80.22 & 88.99 & Beihang University\\
\hline 
A sencce text detection method based on maskrcnn & user\_10680 & 6  & 83.90 & 79.82 & 88.41 &  Fudan University\\
\hline
Fudan-Supremind Detection & user\_9317 & 7  & 83.82 & 81.81 & 85.94 & Fudan University\\
\hline
Amap-CVLab &  user\_4492 & 8  & 83.48 & 81.34 & 85.73 & *\\
\hline
SRCB\_LSVT & user\_1742 & 9  & 83.32 & 77.95 & 89.48 & Samsung Research China-Beijing\\
\hline
HUST\_VLRGROUP &  user\_10862 & 10  & 83.21 & 81.93 & 84.54 & Huazhong University of Science and Technology\\
\hline
pursuer & user\_6849 & 11 & 82.19 & 76.60 & 88.66 & Institute of Automation, Chinese Academy of Sciences\\
\hline
TMIS & user\_9762 & 12 & 81.51 & 78.38 & 84.90 & USTC-iFLYTEK\\
\hline
Mask R-CNN & user\_8368 & 13 & 80.11 & 76.13 & 84.54 & CASIA\\
\hline
PAT-S.Y & user\_9558 & 14 & 79.62 & 73.60 & 86.70 & *\\
\hline
Sg\_whole	& user\_10506 & 15 & 78.69 & 88.23 & 71.01 & Sogou Tech\\
\hline
VIC-LISAR& user\_5607 & 16 & 78.00 & 71.28 & 86.11 & Meituan-Dianping Group\\
\hline
CLTDR & user\_3575 & 17 & 77.52 & 74.94 & 80.28 & Chinese Academy of Sciences\\
\hline
PSENet\_v2 & user\_9422 & 18 & 76.15 & 72.95 & 79.64 & *\\
\hline
Papago OCR (PixelLink+) & user\_10241 & 19 & 73.11 & 67.03 & 80.40 & Papago OCR team, NAVER corp\\
\hline
JDIVA\_Textboxes++	& user\_3058 & 20 & 69.97 & 66.96 & 73.26 & * \\
\hline
test4 & user\_9211 & 21 & 69.95 & 61.75 & 80.66 & Peking University \\ 
\hline
none & user\_7729 & 22 & 68.62 & 62.01 & 76.81& * \\
\hline
one & user\_10042 & 23 & 68.62 & 62.01 & 76.81 & *\\
\hline
CRAFT & user\_8878 & 24 & 66.76 & 64.21 & 69.52 & Clova AI OCR Team, NAVER/LINE Corp\\
\hline
AdvancedEast model with post processing & user\_9059 & 25 & 60.08 & 54.75 & 66.55 & State Key Laboratory of Digital Publishing Technology \& Institute of Computer Science and Technology,Peking University\\
\hline
captcha detection & user\_8033 & 26 & 54.38 & 52.17 & 56.79 & * \\
\hline
TextMask\_V1 & user\_499 & 27 & 45.93 & 77.28 & 32.67 & * \\
\hline
DDT & user\_3442 & 28 & 24.96 & 83.36 & 14.68 & SOGOU\\
\hline
\end{tabular}
\end{table*}

\subsection{TASK 2 - END-TO-END TEXT SPOTTING}
This task aims to detect and recognize text instances from images in an end-to-end manner. Participants are expected to submit the locations of all the text instances in quadrangles or polygons along with the corresponding recognized results.

To compare the end-to-end text spotting results more comprehensively, the submitted models are evaluated in several aspects, including Precision, Recall, H-mean and the normalized metric with Normalized Edit Distance (N.E.D). Under the exactly matched criteria in H-mean, a true positive sample indicates that the Levenshtein distance between the predicted result and the matched ground truth (IoU higher than $0.5$) equals to $0$. The normalized metric $Norm$ equals to $1 - N.E.D$, which is formulated as
\begin{equation}
Norm = 1 - \frac{1}{N}\sum_{i=1}^{N} D(s_i, \hat{s_i})/ \max(l_i, \hat{l_i}),
\end{equation}
\noindent where $D(.)$ stands for the Levenshtein Distance, $s_i$ and $\hat{s_{i}}$ denote the predicted text string and the corresponding ground truth string, and $l_i$, $\hat{l_{i}}$ are their text lengths. Note that the corresponding ground truth $\hat{s_i}$ is calculated over all ground truth locations to select the one in the maximum IoU with the predicted $s_i$ as a pair. $N$ is the maximum number of `paired' ground truths and detected regions, which include singletons, e.g., the ground truth regions that were not matched with any detection (paired with an empty string) and detected regions that were not matched with any ground truth region (paired with an empty string). 

Similar to task 1, \emph{Do not care} regions are excluded and the detected results matched to such regions do not contribute to the final score in edit distance. To avoid the ambiguity in annotations, pre-processing are utilized before comparing two strings: 1) The evaluation for English is case insensitive; 2) The Chinese traditional and simplified characters are considered to be the same categories; 3) The blank spaces and symbols, e.g., comma and dots, etc, are ignored in distance calculation.

\begin{table*}[!hpt]
\caption{End-to-end text spotting results and rankings in Task 2. Note that * denotes missing descriptions in affiliations.}\vspace{-0.5em}
\label{task2_top_10}
\centering
\begin{tabular}
{p{2.4cm}<{\centering}|p{1.4cm}<{\centering}|p{1cm}<{\centering}|p{1.2cm}<{\centering}|p{1cm}<{\centering}|p{0.7cm}<{\centering}|p{1cm}<{\centering}|p{5.5cm}<{\centering}}
\hline
\textbf{Team name} &  \textbf{User ID} & \textbf{Rank} & \textbf{1 - N.E.D} & \textbf{Precision} & \textbf{Recall} & \textbf{H-mean} & \textbf{Affiliations} \\
\hline
\hline
Tencent-DPPR Team &  user\_9062 & 1 & 66.66 & 64.46 & 57.84 & 60.97 & Tencent-DPPR team\\
\hline
HUST\_VLRGROUP & user\_3766 & 2 & 63.42 & 61.75 & 59.39 & 60.55 & Huazhong University of Science and Technology\\
\hline
PMTD & user\_3056 & 3 & 63.36 & 58.67 & 54.97 & 56.76 & SenseTime\\
\hline
\makecell[l]{baseline\_0.7\_polygon \\ \_class\_5435} & user\_9007  & 4 & 63.16 & 59.37 & 53.23 & 56.13 & Beihang University \\
\hline
pursuer & user\_6849 & 5 & 61.51 & 61.31 & 52.96 & 56.83 & Institute of Automation, Chinese Academy of Sciences\\
\hline
MCEM & user\_9762 & 6 & 60.56 & 70.19 & 54.48 & 61.34 & USTC, iFLYTEK\\
\hline
SRC-B\_LSVT & user\_3193 & 7 & 57.48 & 55.40 & 46.83 & 50.76 & Samsung Research China-Beijing \\
\hline
VIC-LISAR & user\_5607 & 8 & 56.00 & 57.00 & 47.13 & 51.60 & Meituan-Dianping Group, Beijing, China\\
\hline
CLTDR & user\_3575 & 9 & 52.94 & 50.57 & 46.11 & 48.24 & Chinese Academy of Sciences\\
\hline
\makecell[l]{Fudan-Supremind \\ Recognition} & user\_6429  & 10 & 52.27 & 47.93 & 39.33 & 43.21 & Fudan University\\
\hline
Papago OCR (PixelLink+) & user\_10241 & 11 & 48.68 & 50.18 & 41.82 & 45.62 & Papago OCR team, NAVER corp\\
\hline
CRAFT + TPS-ResNet v1	& user\_4643 & 12 & 27.59 & 30.10 & 27.48 & 28.73 & Clova AI OCR Team, NAVER/LINE Corp\\
\hline
spotter & user\_8033 & 13 & 26.16 & 21.50 & 19.05 & 20.20 & * \\
\hline
\end{tabular}
\end{table*}

\section{SUBMISSIONS}
There is a total number of $132$ submissions from $41$ teams, and Table. \ref{task1_top_10} and Table. \ref{task2_top_10} summarize all the valid submitted results of the two tasks, respectively.

\subsection{Top 3 submissions in Task 1}
\textbf{Tencent-DPPR team} used Mask R-CNN~\cite{he2017mask} based approach and designed a policy for proposals to select feature pyramid layers to extract features. Multi-scale testing and several trained models including PixelLink~\cite{deng2018pixellink}, Mask R-CNN models with different backbones, i.e., ResNeXt-101~\cite{xie2017aggregated} and ResNet-152~\cite{he2016deep}, were used for model ensembles. ICDAR 2019-LSVT train set was used to train these detectors.\\

\vspace{-0.5em}
\textbf{NJU\_ImagineLab (ImagineLab, National Key Lab for Novel Software Technology)} followed the Mask R-CNN framework, utilizing ResNet-152 as the backbone, deformable ConvNets v2~\cite{zhu2018deformable} in the last three stages and PANet~(Path Aggregation Network)~\cite{liu2018path} to fuse multi-scale features. During the training stage, OHEM~(Online Hard Example Mining)~\cite{dai2016r} was utilized to sample the hard proposals from ICDAR 2019-LSVT train set. In the testing stage, multi-scale testing was used to get the final results.\\

\vspace{-0.5em}
\textbf{PMTD (SenseTime)} utilized PMTD~(Pyramid Mask Text Detector)~\cite{2019pmtd} as the detection method based on Mask-RCNN with several modifications. During the training stage, they utilized the ICDAR 2019-LSVT train set to learn the text detector. During the testing stage, multi-scale testing was also adopted.

\subsection{Top 3 submissions in Task 2}
\textbf{Tencent-DPPR team} used a two-stage detection method as described in task 1. For text recognition, CNN was used to extract text features for sequence decoding. Different text recognition models were separately trained on horizontal and vertical text instances, e.g., bi-directional GRU~(Bi-directional Gated Recurrent Unit) with CTC~(Connectionist Temporal Classification )~\cite{vaswani2017attention} loss for transcription,  attention-based RNN decoding~\cite{bahdanau2014neural} and multi-head self-attention~\cite{vaswani2017attention} encoder with CTC loss for decoding. For model ensembles, different iterations of each type of models were used to generate candidate results and select the most reliable one according to the prediction confidence.  The training data included a synthetic dataset with more than $50$ million images, as well as publicly released datasets, i.e., ICDAR 2019-LSVT, ReCTS, ICDAR 2017 COCO-Text, ICDAR 2017-RCTW and ICPR 2018-MTWI.



\textbf{HUST\_VLRGROUP} adopted the Mask TextSpotter~\cite{he2018end} based method and used the ResNet-50-FPN as the backbone network. Deformable convolutions~\cite{dai2017deformable} were used in the last two stages to enhance features. For text detection, Cascade R-CNN~\cite{cai2018cascade} was used to generate text proposals. For the recognition part, CNN and RNN with CTC loss~\cite{shi2017end} was the adopted and several models were trained with multiple backbones, including ResNext-50 and inception modules to obtain the final results. To handle irregular text instances, a rectification module with STN~(Spatial Transformer Network)~\cite{jaderberg2015spatial} was developed as pre-processing layers before recognition. The training data included the publicly available datasets, i.e., ICDAR 2019-LSVT, ICDAR 2003, ICDAR 2013, ICDAR 2015, IIIT5K, ICDAR 2017 COCO-Text, ICDAR 2017-MLT English + Chinese, ICDAR 2017 RCTW-17, ICPR 2018-MTWI, CTW and synthetic images.
\textbf{PMTD (SenseTime)} generated detection results with PMTD, and utilized the recognition method based on CNN and CTC. 
Training data were collected from ICDAR 2019-LSVT, ICPR 2018-MTWI and ReCTS. During testing stage, every detected text region was recognized without additional post-processing.\\

\section{ANALYSIS}
\begin{figure}[ht]
	\begin{center}
		\includegraphics[width=\linewidth]{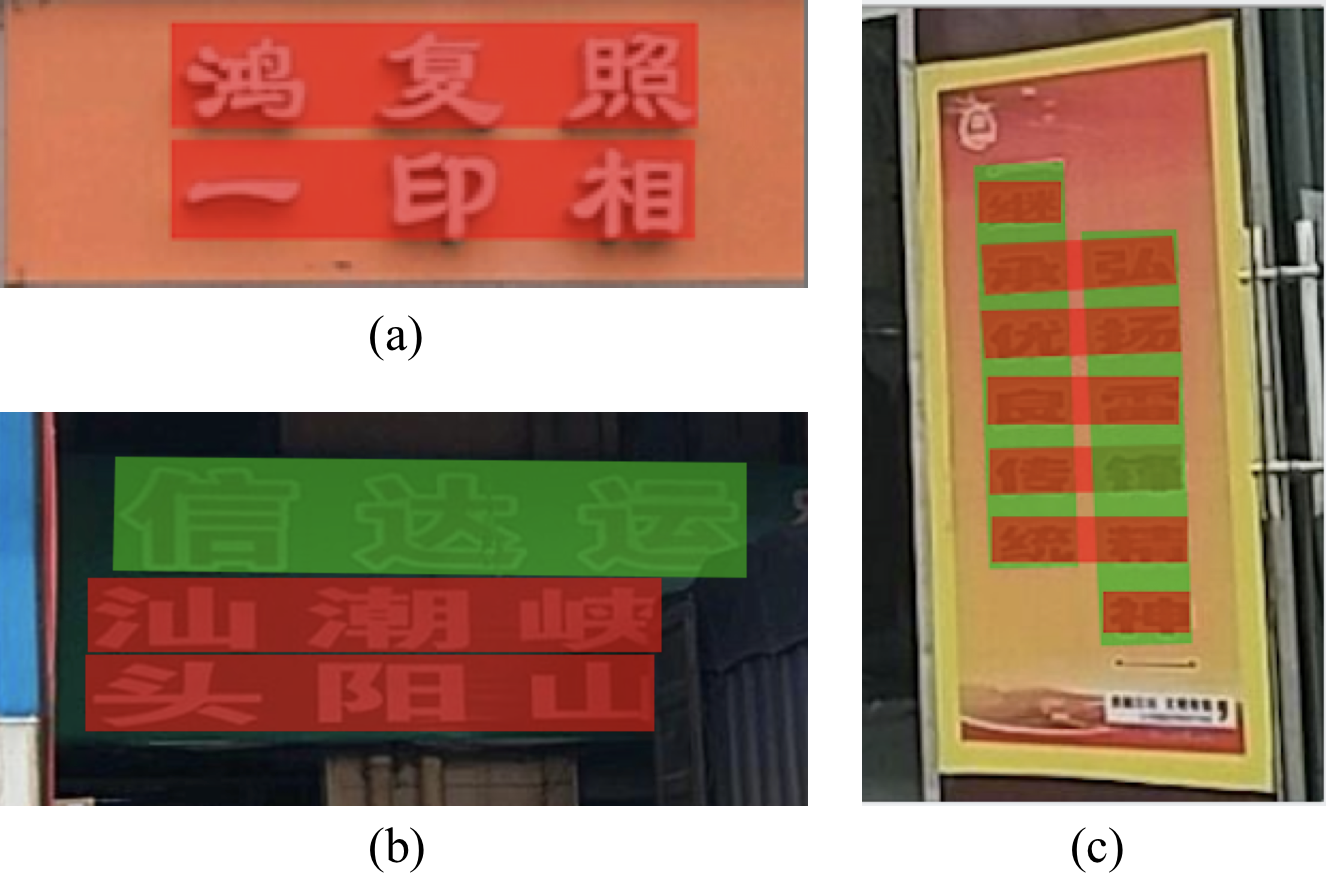}
	\end{center}\vspace{-1.5em}
	\caption{Examples of bad cases for text detection in the competition cropped from the test set of ICDAR 2019-LSVT. Note that the green masks indicate correctly localized regions, while red masks are labeled as bad cases in detection errors.} 
	\label{fig:analysis_detect}
\end{figure}

Most participants in Task.1 employ instance-level segmentation methods to accurately localize text instances, following the framework of Mask R-CNN. Backbone features, e.g., ResNeXt, deformable convolutional layers and PANet, are used to further improve the localization results for long and curved text instances. The Tencent-DPPR team achieves the best score in H-mean, recall and precision with enhanced backbones and model ensembles. The visualization analysis shows that one of the common bad cases include the mis-detection and grouping errors of vertical text lines. Most of the trained models intend to group them as horizontal lines as shown in Fig.~\ref{fig:analysis_detect} (a), (b) and (c), respectively. In this case, it may be useful to exploit the semantic information to decide how to group or split text instances.

\begin{figure}[t]
\centering
\subfigure[Large variations in fonts and styles]{\includegraphics[width=\linewidth]{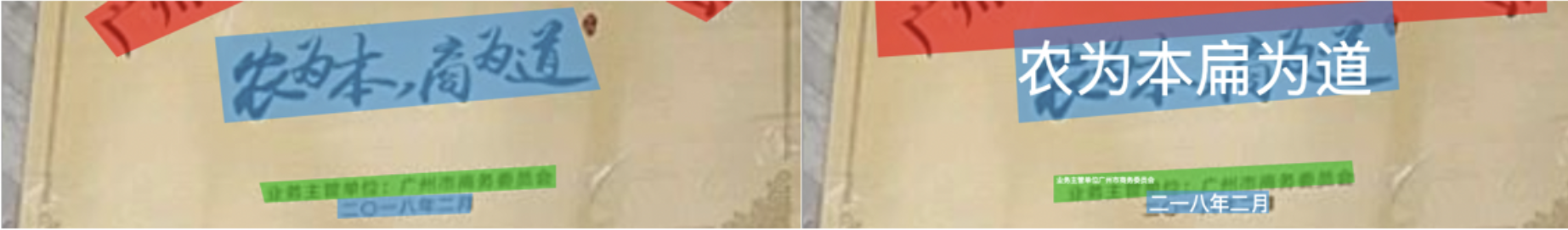}}
\subfigure[Rotations of text instances] {\includegraphics[width=\linewidth]{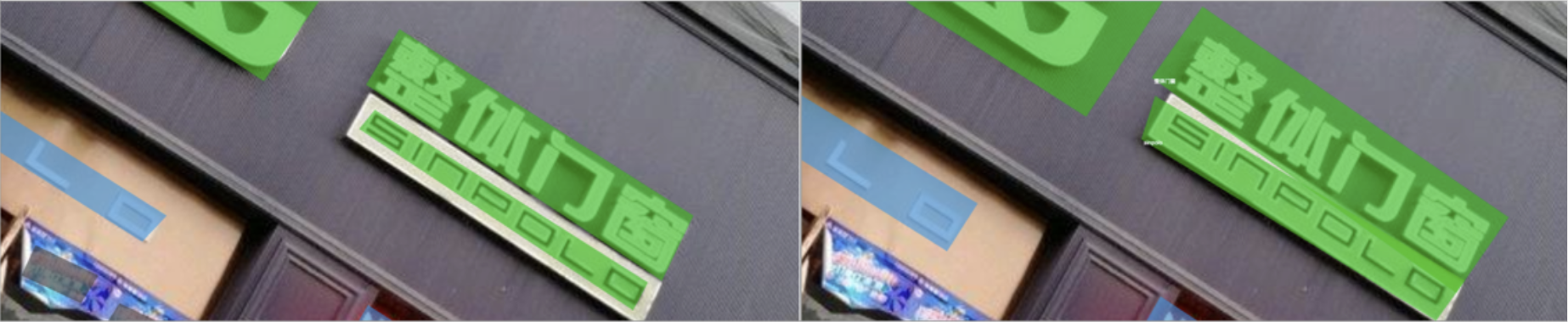}}
\subfigure[Characters placed in low contrast with complex backgrounds]{\includegraphics[width=\linewidth]{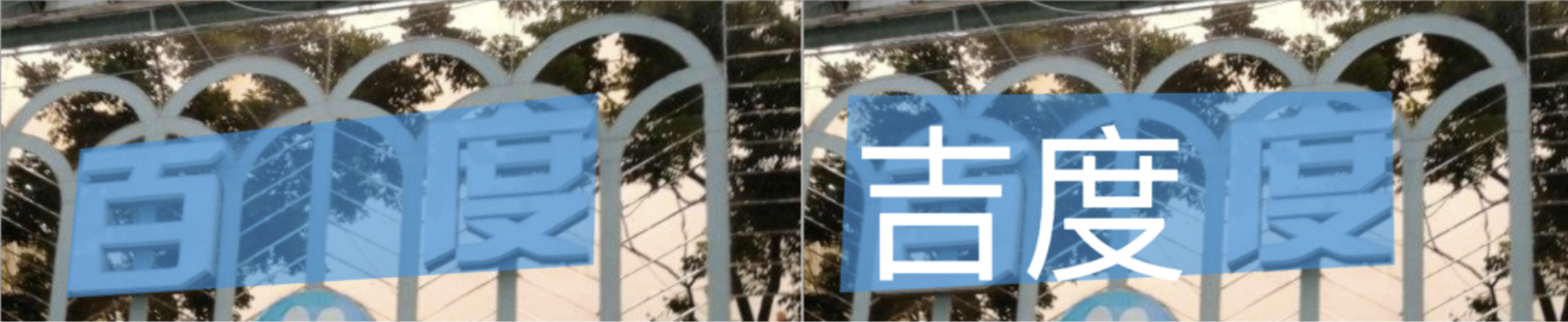}}
\subfigure[Partial occlusions of a few characters]{\includegraphics[width=\linewidth]{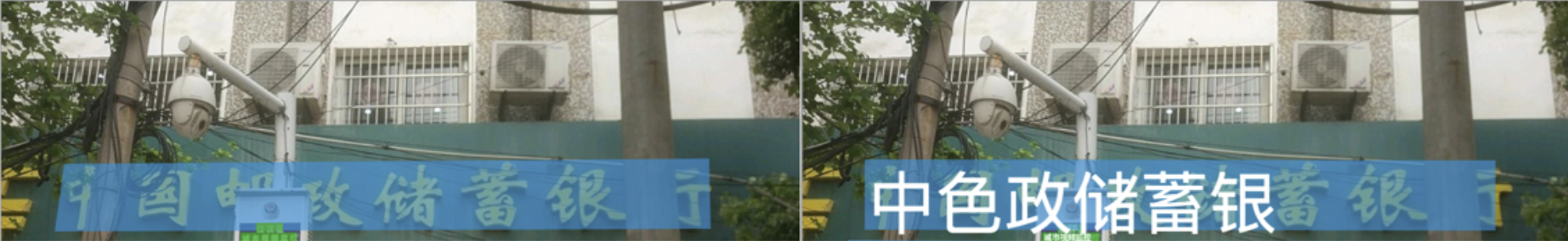}}\vspace{-0.5em}
\caption{Examples of text recognition results in the competition cropped from the test set of ICDAR 2019-LSVT. The green mask indicates that the regions are detected and recognized correctly, while the blue mask indicates that there exist text recognition errors in the results (best view in colors).}
	\label{fig:analysis_recog}
\end{figure}

For text spotting task, most of the submitted methods adopt a two-stage based pipeline for end-to-end recognition, considering detection and recognition as two modules. In recognition part, most participants employ CNN and Bi-LSTM/Bi-GRU encoders with CTC loss for prediction and training. Tencent DPPR team ranks the first in terms of precision, H-mean and Norm by model ensembles, including CNN + Bi-GRU + CTC, CNN + Bi-GRU + Attention based sequence-to-sequence and CNN + Self-Attention + CTC models. The HUST team in the second place gets the top performance in recall and a rectification module with STN in recognition is used to handle irregular text instances. In comparisons with existing scene text benchmarks, text instances from real-world street views have larger variations in font, size, aspect ratios, orientations and arbitrary shapes as the remaining challenges to solve as shown in the Fig.~\ref{fig:analysis_recog} (a) and (b), respectively. 

The mentioned two-stage pipeline has not made full use of the highly relevant and complementary relationship between detection and recognition. The performance of end-to-end text recognition highly relies on the recall and precision of text detection results. The feedback of recognition module is helpful to improve text detection results. Bad cases in common indicate the lack of robustness to complex backgrounds, e.g., low contrast in luminance and partial occlusions of a few characters as shown in Fig.~\ref{fig:analysis_recog} (c) and (d), respectively. To further improve the recognition performance with better invariance, it is beneficial to make full use of large-scale datasets with rich context information in full and weak annotated data.


\section{CONCLUSIONS AND FUTURE DIRECTIONS}
This paper summarizes the organization and results of ICDAR 2019-LSVT challenge in the RRC series. Large-scale street view text dataset was collected and annotated in full and weak annotations, respectively, making it the largest ICDAR scene text dataset ever. There have been a number of $41$ teams participating in the proposed two tasks and $132$ valid submissions in total, which have shown great interest from both research communities and industries. Results analysis has shown the abilities of state-of-the-art text detection and recognition systems, summarizing the difficulties, remaining challenges and future research directions. All the details about ICDAR 2019-LSVT and datasets are available on the RRC websites. 

In the future, we intend to keep on maintaining the ICDAR 2019-LSVT competition leaderboard to encourage more participants to submit and improve their results, which aims to help bridge the gap between research and industrial applications and build a smarter text reading system approaching human intelligence.

\section*{Acknowledgment}
The ICDAR 2019-LSVT competition is sponsored by Baidu Inc., Beijing, China. The organizers would like to thank all the participants for their valuable and helpful feedback, which has also contributed to the success of the competitions.



\bibliographystyle{IEEEtran}
\bibliography{baidu_references,egbib}

\begin{thebibliography}{10}
\providecommand{\url}[1]{#1}
\csname url@samestyle\endcsname
\providecommand{\newblock}{\relax}
\providecommand{\bibinfo}[2]{#2}
\providecommand{\BIBentrySTDinterwordspacing}{\spaceskip=0pt\relax}
\providecommand{\BIBentryALTinterwordstretchfactor}{4}
\providecommand{\BIBentryALTinterwordspacing}{\spaceskip=\fontdimen2\font plus
\BIBentryALTinterwordstretchfactor\fontdimen3\font minus
  \fontdimen4\font\relax}
\providecommand{\BIBforeignlanguage}[2]{{%
\expandafter\ifx\csname l@#1\endcsname\relax
\typeout{** WARNING: IEEEtran.bst: No hyphenation pattern has been}%
\typeout{** loaded for the language `#1'. Using the pattern for}%
\typeout{** the default language instead.}%
\else
\language=\csname l@#1\endcsname
\fi
#2}}
\providecommand{\BIBdecl}{\relax}
\BIBdecl

\bibitem{yao2012detecting}
C.~Yao, X.~Bai, W.~Liu, Y.~Ma, and Z.~Tu, ``Detecting texts of arbitrary
  orientations in natural images,'' in \emph{IEEE Conference on Computer Vision
  and Pattern Recognition}, 2012, pp. 1083--1090.

\bibitem{karatzas13icdar}
D.~Karatzas and et~al, ``{ICDAR} 2013 robust reading competition,'' in
  \emph{Proc. of ICDAR}.\hskip 1em plus 0.5em minus 0.4em\relax IEEE, 2013, pp.
  1484--1493.

\bibitem{karatzas15icdar}
------, ``{ICDAR} 2015 competition on robust reading,'' in \emph{Proc. of
  ICDAR}.\hskip 1em plus 0.5em minus 0.4em\relax IEEE, 2015, pp. 1156--1160.

\bibitem{liu2011casia}
C.-L. Liu, F.~Yin, D.-H. Wang, and Q.-F. Wang, ``{CASIA} online and offline
  chinese handwriting databases,'' in \emph{2011 International Conference on
  Document Analysis and Recognition (ICDAR)}.\hskip 1em plus 0.5em minus
  0.4em\relax IEEE, 2011, pp. 37--41.

\bibitem{icdar2017-mlt}
``{ICDAR} 2017 competition on multilingual scene text detection and script
  identification,'' \url{http://rrc.cvc.uab.es/?ch=8&com=introduction},
  accessed: 2018-11-16.

\bibitem{chng17tt}
C.~K. Chng and C.~S. Chan, ``Total-text: {A} comprehensive dataset for scene
  text detection and recognition,'' in \emph{Proc. of ICDAR}, 2017.

\bibitem{yuliang2017detecting}
L.~Yuliang, J.~Lianwen, Z.~Shuaitao, and Z.~Sheng, ``Detecting curve text in
  the wild: New dataset and new solution,'' \emph{arXiv preprint
  arXiv:1712.02170}, 2017.

\bibitem{shi2017icdar2017}
B.~Shi, C.~Yao, M.~Liao, M.~Yang, P.~Xu, L.~Cui, S.~Belongie, S.~Lu, and
  X.~Bai, ``{ICDAR2017} competition on reading chinese text in the wild
  ({RCTW}-17),'' in \emph{2017 14th IAPR International Conference on Document
  Analysis and Recognition (ICDAR)}, vol.~1.\hskip 1em plus 0.5em minus
  0.4em\relax IEEE, 2017, pp. 1429--1434.

\bibitem{yuan2018chinese}
T.-L. Yuan, Z.~Zhu, K.~Xu, C.-J. Li, and S.-M. Hu, ``Chinese text in the
  wild,'' \emph{arXiv preprint arXiv:1803.00085}, 2018.

\bibitem{jaderberg2014deep}
M.~Jaderberg, A.~Vedaldi, and A.~Zisserman, ``Deep features for text
  spotting,'' in \emph{ECCV}, 2014.

\bibitem{he2017deep}
W.~He, X.-Y. Zhang, F.~Yin, and C.-L. Liu, ``Deep direct regression for
  multi-oriented scene text detection,'' in \emph{ICCV}, 2017.

\bibitem{xiangcvpr2017}
B.~Shi, X.~Bai, and S.~Belongie, ``Detecting oriented text in natural images by
  linking segments,'' in \emph{CVPR}, 2017.

\bibitem{TextBoxes}
M.~Liao, B.~Shi, X.~Bai, X.~Wang, and W.~Liu, ``Textboxes: {A} fast text
  detector with a single deep neural network,'' in \emph{AAAI}, 2017.

\bibitem{DMPNet}
L.~Liu, Yuliang~Jin, ``Deep matching prior network: Toward tighter
  multi-oriented text detection,'' in \emph{CVPR}, 2017.

\bibitem{Liao2018TextBoxesAS}
M.~Liao, B.~Shi, and X.~Bai, ``{TextBoxes++}: A single-shot oriented scene text
  detector,'' \emph{T-IP}, 2018 (in Press).

\bibitem{Gupta16}
A.~Gupta, A.~Vedaldi, and A.~Zisserman, ``Synthetic data for text localisation
  in natural images,'' in \emph{IEEE Conference on Computer Vision and Pattern
  Recognition}, 2016.

\bibitem{veit2016coco}
A.~Veit, T.~Matera, L.~Neumann, J.~Matas, and S.~Belongie, ``Coco-text: Dataset
  and benchmark for text detection and recognition in natural images,''
  \emph{arXiv preprint arXiv:1601.07140}, 2016.

\bibitem{he2017mask}
K.~He, G.~Gkioxari, P.~Doll{\'a}r, and R.~Girshick, ``Mask r-cnn,'' in
  \emph{Proceedings of the IEEE international conference on computer vision},
  2017, pp. 2961--2969.

\bibitem{deng2018pixellink}
D.~Deng, H.~Liu, X.~Li, and D.~Cai, ``Pixellink: Detecting scene text via
  instance segmentation,'' in \emph{Thirty-Second AAAI Conference on Artificial
  Intelligence}, 2018.

\bibitem{xie2017aggregated}
S.~Xie, R.~Girshick, P.~Doll{\'a}r, Z.~Tu, and K.~He, ``Aggregated residual
  transformations for deep neural networks,'' in \emph{Proceedings of the IEEE
  conference on computer vision and pattern recognition}, 2017, pp. 1492--1500.

\bibitem{he2016deep}
K.~He, X.~Zhang, S.~Ren, and J.~Sun, ``Deep residual learning for image
  recognition,'' in \emph{Proceedings of the IEEE conference on computer vision
  and pattern recognition}, 2016, pp. 770--778.

\bibitem{zhu2018deformable}
X.~Zhu, H.~Hu, S.~Lin, and J.~Dai, ``Deformable convnets v2: More deformable,
  better results,'' \emph{arXiv preprint arXiv:1811.11168}, 2018.

\bibitem{liu2018path}
S.~Liu, L.~Qi, H.~Qin, J.~Shi, and J.~Jia, ``Path aggregation network for
  instance segmentation,'' in \emph{Proceedings of the IEEE Conference on
  Computer Vision and Pattern Recognition}, 2018, pp. 8759--8768.

\bibitem{dai2016r}
J.~Dai, Y.~Li, K.~He, and J.~Sun, ``R-fcn: Object detection via region-based
  fully convolutional networks,'' in \emph{Advances in neural information
  processing systems}, 2016, pp. 379--387.

\bibitem{2019pmtd}
J.~S. D. L. X. L. Q.~L. Jingchao~Liu, Xuebo~Liu, ``Pyramid mask text
  detector,'' \emph{arXiv preprint arXiv:1903.11800}, 2019.

\bibitem{vaswani2017attention}
A.~Vaswani, N.~Shazeer, N.~Parmar, J.~Uszkoreit, L.~Jones, A.~N. Gomez,
  {\L}.~Kaiser, and I.~Polosukhin, ``Attention is all you need,'' in
  \emph{Advances in neural information processing systems}, 2017, pp.
  5998--6008.

\bibitem{bahdanau2014neural}
D.~Bahdanau, K.~Cho, and Y.~Bengio, ``Neural machine translation by jointly
  learning to align and translate,'' \emph{arXiv preprint arXiv:1409.0473},
  2014.

\bibitem{he2018end}
T.~He, Z.~Tian, W.~Huang, C.~Shen, Y.~Qiao, and C.~Sun, ``An end-to-end
  textspotter with explicit alignment and attention,'' in \emph{Proceedings of
  the IEEE Conference on Computer Vision and Pattern Recognition}, 2018, pp.
  5020--5029.

\bibitem{dai2017deformable}
J.~Dai, H.~Qi, Y.~Xiong, Y.~Li, G.~Zhang, H.~Hu, and Y.~Wei, ``Deformable
  convolutional networks,'' in \emph{Proceedings of the IEEE international
  conference on computer vision}, 2017, pp. 764--773.

\bibitem{cai2018cascade}
Z.~Cai and N.~Vasconcelos, ``Cascade r-cnn: Delving into high quality object
  detection,'' in \emph{Proceedings of the IEEE Conference on Computer Vision
  and Pattern Recognition}, 2018, pp. 6154--6162.

\bibitem{shi2017end}
B.~Shi, X.~Bai, and C.~Yao, ``An end-to-end trainable neural network for
  image-based sequence recognition and its application to scene text
  recognition,'' \emph{IEEE transactions on pattern analysis and machine
  intelligence}, vol.~39, no.~11, pp. 2298--2304, 2017.

\bibitem{jaderberg2015spatial}
M.~Jaderberg, K.~Simonyan, A.~Zisserman \emph{et~al.}, ``Spatial transformer
  networks,'' in \emph{Advances in neural information processing systems},
  2015, pp. 2017--2025.

\end{thebibliography}

%

\end{document}